# AN IMPLEMENTATION OF APERTIUM BASED ASSAMESE MORPHOLOGICAL ANALYZER

Mirzanur Rahman[1] and Shikhar Kumar Sarma[2]

Department of Information Technology, Gauhati University, Guwahati, Assam, India

## ABSTRACT

*Morphological Analysis is an important branch of linguistics for any Natural Language Processing Technology. Morphology studies the word structure and formation of word of a language. In current scenario of NLP research, morphological analysis techniques have become more popular day by day. For processing any language, morphology of the word should be first analyzed. Assamese language contains very complex morphological structure. In our work we have used Apertium based Finite-State-Transducers for developing morphological analyzer for Assamese Language with some limited domain and we get 72.7% accuracy*

## 1. INTRODUCTION

Assamese is the major language spoken in Assam. The state Assam is the north –eastern part of the country. Assamese Language served as a bridge language among different speech communities in the whole area of the state. The language Assamese is an Indo-Aryan language originated from Vedic dialects [1]. The language as it stands today, passes through tremendous modifications in all the component viz. phonology, morphology, conjunction etc. There are two variations of Assamese language according to dialectical regions [2] i.e. Eastern Assamese and Western Assamese Language. Both are different in terms of phonology and morphology. But still the written text is same for all the regions.

Morphology is an important component of any language. So, before any processing, we must first analyze the morphology of the words of that language. In language processing technology such as Machine Translation [15], Parsing, POS Tagger, Text summarization etc requires morphological analyzers to find out the lexical component of a word. And lexical components are the very important parts of a grammar of a language.

In our work we have considered only the standard written Assamese Text corpus and use Apertium engine with Lttoolbox to build morphological analyzer for Assamese language. In this paper we have discuss how we proceed towards developing morphological analyzer.

## 2. MORPHOLOGY OF A LANGUAGE

In Language Technology research, morphological analysis studies the structure of words and word formation of a language.

     

Words in a language can be divided into many small units which are known as morpheme [10]. Recognizing different morpheme in a word with their lexical properties is known as morphological analysis. For example in English language

- **Girls=Girl +s**
    - Root : Girl (category- noun)
    - Affix: 's' (indefinite plural marker)

In the above example the word "Girls" is a combination of "Girl" morpheme and "s" morpheme. When we analyze a word with morphological analyzer, it should provide all the combination of morpheme with their lexical properties.

According to Golockchandra Goswami [1], all the morphemes of Assamese Language can be divided in to three categories.

- Root Morpheme
- Sub- Root Morpheme
- Affixes Morpheme

Root morphemes are the main morpheme depending on which all the morphological construction is done and affixes are attached. Root morpheme may be categorize into other Parts-of-speech category like noun, pronoun, verb, indeclinable etc.

For example in Assamese Language:

- ল'ৰাজন = ল'ৰা + জন
    - Root: ল'ৰা (Category -Noun)
    - Affix: জন (Definite Article, singular case marker)

Sub-root morphemes are the morpheme which can occur as an independent root as well as a suffix in an Assamese sentence.

For example:

- খন, জন ডাল etc. can be used as a root word or as a suffix for singular definitive
    - মানুহজন বৰ ভাল ("জন" is used as a singular suffix marker)
    - মোৰ জন আহিলেনে? ("জন" is used as a independent root)
- বোৰ/বোৰৰ, বিলাক etc can be used as a root word or as a suffix for plural definitive
    - বস্তুবোৰৰ দাম নাই। ("বোৰৰ" is used as a plural suffix marker)
    - তোমালোকৰ বোৰৰ কি খবৰ? ("বোৰৰ" is used as a independent root)

Affixes are always added to the root and it contains some own meaning. But they have no separate existence and can never form a free form alone or in conjunction with themselves.

For Example: এ (মানুহে), ক (মানুহক), লৈ (তোমালৈ),কৈ (ভালকৈ) etc

## 3. PRIOR ARTS

For Assamese language also we have found some of the reported work for morphological analysis. In this section we will try to summarize all the reported work related to Assamese morphological analysis.

- In [3], the authors have presented building Morphological Analyzers using the Suffix Stripping method for the four languages – Assamese, Bengali, Bodo and Oriya. In the proposed mechanism they have deals with only inflectional suffixes. The method involves identifying individual suffixes from a series of suffixes attached to a stem/root, using morpheme sequencing rules.

  In the approach the analyzer analyses the inflected form of a word into suffixes and stems by using a root/stem dictionary (for identifying legitimate roots/stems), a list of suffixes, comprising of all possible suffixes that various categories can take (in order to identify a valid suffix), and the morpheme sequencing rules. . The authors get 50 % coverage for 7000 to 8000 root entries.

- In [4], the authors have presented A Suffix-based Noun and Verb Classifier for an Inflectional Language. In the proposed mechanism they have consider only the morpho-syntactic properties of Assamese words. Assamese words can be categorized into inflected classes (noun, pronoun, adjective and verb) and un-inflected classes (adverb and particle.

- In [5], the authors describe an approach to unsupervised learning of morphology from an unannotated corpus for Assamese Language in their paper "Acquisition of Morphology of an Indic Language from Text Corpus". In their paper they have present & elaborately discussed an unsupervised method for acquisition of Assamese morphology from a text corpus. This is the initial work towards unsupervised morphological analysis and it is very suitable for Assamese language. This approach, acquire the suffixation morphology of the language from a text corpus of about 300,000 words and build a morphological lexicon. The F-measure of the suffix acquisition is about 69%.

- In [6], the authors have presented suffix stripping approach, where they add a rule engine which generates all the possible suffix sequences for analyzing morphology of a word. They got 82% accuracy with a root-word list of size 20,000 approximately with this method.

- In [7], the authors combine a rule based algorithm and HMM based algorithm. Where rule based algorithm is used for predicting multiple letter suffixes and an HMM based algorithm for predicting the single letter suffixes .This added method can predict morphologically inflected words with 92% accuracy.

- In [8] Utpal Sarma proposed an unsupervised method for learning morphology of a language in his Ph.D thesis "Unsupervised Learning of Morphology of a Highly Inflectional Language"

## 4. IMPLEMENTATION USING APERTIUM AND LTTOOLBOX

Apertium is a rule-based open-source shallow-transfer machine translation platform [11]. It is free software and released under the terms of the GNU General Public License. It includes the engine, maintenance tools, and open linguistic data for several language pairs. Lttoolbox is a toolbox for lexical processing, morphological analysis and generation of words. Lttoolbox used finite-state transducers (FST). FST are a type of finite-state automata, which may be used as one-pass morphological analyzers.

In Apertium, the analyzer data is stored in Apertium's dictionary (dix) format with XML syntax. The analyzer can be easily converted to a morphological generator from the single morphological database (monodix), depending on in which direction the system read the dictionary. If the system read the dictionary from left to right, we obtain the analyzer, and read from right to left, we obtain the generator. It is proven that an XML based dictionary (monodix) is generally faster than a normal text or database based dictionary.

For creating Morphological Analyzer, different modules of Apertium engine are required.

### 4.1. Dictionary

An Apertium based system can use two types of dictionaries, Monolingual and Bilingual Dictionary. Monolingual dictionary is used for Morphological analyzer & generator and Bilingual dictionary is used for machine translation purpose. In our work, we use monolingual dictionary.

### 4.2. Paradigm definitions <pardef>

A Paradigm is the complete set of related inflectional and productive derivational word forms of a given category. A paradigm can be understood as a small dictionary of alternative transformations that can be concatenated to the parts of words (or to entries of another paradigm) to specify regularities in the lexical processing of the dictionary entries, such as inflection regularities. In the definition along with the root word it contains other information like category, gender, number, person, case marker, tense etc.

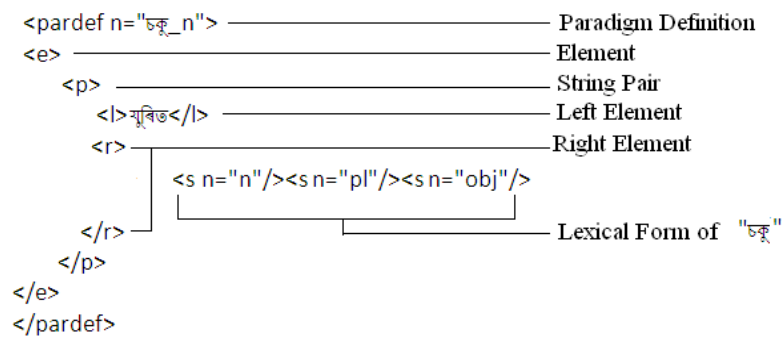

Figure 1: Dictionary entry of Assamese Word চকু (Eye) in Paradigm

### 4.3. Element for Reference to a Paradigm

Apertium provides a lexico-semantic layer, for working with inflection of a word. The layer introduces the lexemes into derivation and concurrently follows the inflection of the derived lexeme. It is used inside <pardefs> entry. Main advantage of using reference paradigm is that, there is no need to write all the inflected forms of a lemma in a morphological dictionary entry because it can be referred from other paradigms.

```
<e lm="চকু">
    <i>চকু</i>
    <par n="চকু_n"/>  ——— Reference Paradigms for Inflection
</e>
```

Figure 2: XML format for Reference Paradigm

### 4.4. Morpheme

All the root word (morpheme) is included in the dictionary, generated for Morphological Analyzer. The dictionary is different from a conventional dictionary, because it contains other information with morpheme like lexical categories and their corresponding paradigm.

```
                                    ——— Morpheme
<pardef n="মানুহ_n">
    <e>
      <p>
        <l>মানুহ</l>
        <r><s n="n"/><s n="sg"/></r>  ——— Lexical Category of Morpheme
      </p>
    </e>
</pardef>
```

Figure 3: Dictionary Entry of Morpheme with their lexical category

### 4.5. Lttoolbox Modules

Lttoolbox contains three modules, lexical processing (lt-comp), morphological analysis & generation (lt-proc) and Expansion (lt-expand).

For Morphological analysis lt-comp and lt-proc module is required, lt-comp for processing and lt-proc for generation [12]. Lt-comp module is responsible for compiling our morphological dictionaries into its own finite-sate representation and lt-comp module is responsible for processing the compiled input data into required output.

4.5.1. **Compilation:**

lt-comp module compile the given .dix format file into binary format from left to right (LR) or from right to lest (RL). When we compile with LR, it creates an analyzer and RL usually creates a generator.

**Syntax of lt-comp:**
$ lt-comp  lr  apertium-asm.morph.dix  analyser.bin
Compile the apertium-asm.morph.dix  dictionary in a left-to-right manner into the binary analyser.bin
$ lt-comp rl  apertium-asm.morph.dix  generator.bin

Compile the apertium-asm.morph.dix dictionary in a right-to-left manner into the binary generator.bin

4.5.2. **Processing:**

lt-proc module contains two functions, one is analysis (which is the default mode) and generation. Analysis converts surface forms into the set of possible lexical forms, while generation converts a lexical form into the corresponding surface form.

**Syntax of lt-proc:**

$ echo "চকুযুৰি" | lt-proc analyser.bin
Output:      ^চকুযুৰি/চকু<n><pl>$

Here we analyze the Assamese word চকুযুৰি (Eyes) with the binary format dictionary (left-to-right) analyser.in

 $ echo "^চকু<n><pl>$" | lt-proc -g generator.bin
 Output:    চকুযুৰি

Here we generate the plural form of Assamese word চকু (Eye) with the binary format dictionary (right-to-left) generator.bin

## 4.6. Meaning of Analyzers Output format

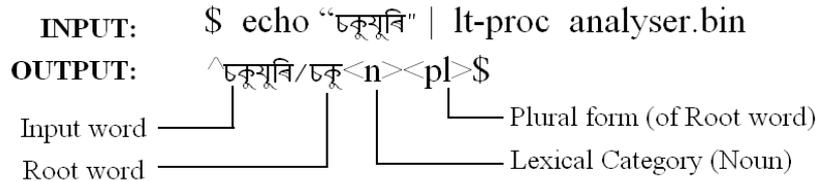

Figure 4: Meaning of output format (Morphological Analyzer)

## 4.7. Current Dictionary

In our current work, we have considered only limited number of word with selected Parts-of-Speech (POS) categories. The following table shows summery of our used database entries

Table 1: Number of Entries in XML Dictionary (apertium-asm.morph.dix)

| Main Category | Entry in dictionary |
|---|---|
| Noun | 22368 |
| Pronoun | 121 |
| Verb | 1844 |
| Adverb | 232 |

As a source of our dictionary, we have use following sources

- Assamese text corpus obtained from Language Technology Development Project, Gauhati University
- Asamiya Abhidhan [9]

All the dictionary entries are done manually by using notepad++ software. The root words lexical categories are verified by linguistics, so that the output of the analyzer is proper.

## 5. EVALUATION AND TEST RESULT

In the above table (Table no 1) we have seen that the number of dictionary entry is only 24565, which is not a very high in number. Since the work is going on, we can expect in future we will have an XML dictionary with large number of entries. Here we have considered only the most frequently used words in Assamese Language.

Till now we have not added any rule for lexical selection in the Apertium engine. That's why some times it cannot analyze a word properly. For example, the Assamese word জন can be used as a suffix or it can be used as a person name. Most of the time the word জন (man in general) (Definite Article) is used as a suffix in Assamese sentences, but if someone use this word as a person name (Though জন(jhon) (Proper noun) is not commonly used as a name in Assam) then our analyzer cannot give proper analysis. We have faced another problem with this analyzer is that if a word has more than one meaning depending on the situation and position within the sentence, it cannot analyze properly. For Example the Assamese Words: মালা (Garland) and অনল (Fire). Both can be used as a material noun or proper noun depending on the use of the word in the sentence.

In testing phase we have use a set of data collected from different Assamese blogs and pages containing 1120 words (after Cleaning). Words are first tokenize and passes through cleaning process (for removing stop word, delimiter and extra white space ) with the help of java programming language. Then one by one we pass the word to the Apertium engine for analysis and store the result in a text file. The text file is checked manually for correctness of the results. The result we have found is shown below

Table 2: Test Result

| Total words | 1120 |
|---|---|
| Correctly recognize | 815 |
| Wrongly recognize | 305 |

From the above table we have seen that the analyzer provides only 72.7% correct results .Other 27% are wrongly recognize due to limited database entry , unavailability of lexical rules for selecting proper category and limited POS category.

## 6. CONCLUSION

In this paper we have discus about the implementation of a Morphological analyzer using Apertium & Lttoolbox. At present this analyzer can handles only inflectional morphology, since we are excluding derivational morphology and we are working on noun, pronoun, verb and adverb. Our current dictionary can only provide information about suffixes.

Form the previous works (in section III) we can see that maximum works done with supervised suffix stripping method. Only limited no of [5, 8] reported work has implement unsupervised technique for analyzing the morphology. Here we have used supervised Finite-state-transducer (FST) method with the help of Apertium engine, since Finite-state-transducers have many advantages [13]. With the help of single source, FST can work as bidirectional engine for both analysis and generation and they are fast (thousands of words per second), and compact.

Currently our morphological analyzer is in initial stage. In the future we will extend our work to the remaining grammatical categories, include derivational morphology and populate dictionary with prefix information to get better performance.

## ACKNOWLEDGEMENTS

The authors are thankful to the Department of Information Technology, Gauhati University for providing us the corpus, which helped us in building the MA system and people from Language Technology Development Project, Gauhati University for their immense support

**Authors**

**Mirzanur Rahman:** PhD Scholar, Department of Information Technology, Gauhati University.

**Shikhar Kr. Sarma:** Head of the Department, Department of Information Technology, Gauhati University.